\documentclass{article} 
\usepackage{graphicx}
\usepackage{hyperref}
\usepackage{amssymb}
\usepackage{float}
\usepackage{wrapfig}
\usepackage{subcaption}
\usepackage[parfill]{parskip}

\title{Mathematical Reasoning in Latent Space}
\date{September 2019}

\author{Dennis Lee, Christian Szegedy, Markus N. Rabe, Sarah M. Loos and Kshitij Bansal \\
Google Research\\
Mountain View, CA, USA\\
\texttt{\{ldennis,szegedy,mrabe,smloos,kbk\}@google.com}}

\begin{document}
\maketitle
\begin{abstract}

We design and conduct a simple experiment to study whether neural networks can perform several steps of approximate reasoning in a fixed dimensional latent space.
The set of rewrites (i.e. transformations) that can be successfully performed on a statement represents essential semantic features of the statement.
We can compress this information by embedding the formula in a vector space, such that the vector associated with a statement can be used to predict whether a statement can be rewritten by other theorems.
Predicting the embedding of a formula generated by some rewrite rule is naturally viewed as approximate reasoning in the latent space.
In order to measure the effectiveness of this reasoning, we perform approximate deduction sequences in the latent space and use the resulting embedding to inform the semantic features of the corresponding formal statement (which is obtained by performing the corresponding rewrite sequence using real formulas).
Our experiments show that graph neural networks can make non-trivial predictions about the rewrite-success of statements, even when they propagate predicted latent representations for several steps. Since our corpus of mathematical formulas includes a wide variety of mathematical disciplines, this experiment is a strong indicator for the feasibility of deduction in latent space in general.
\end{abstract}

\section{Introduction}

Automated reasoning has long been considered to require development of logics and ``hard'' algorithms, such as backtracking search. 
Recently, approaches that employ deep learning have also been applied, but these have focused on predicting the next step of a proof, which is again executed with a hard algorithm~\cite{loos2017deep,gauthier2017tactictoe,Lederman2018QBF,bansal2019holist}.

We raise the question whether hard algorithms could be omitted from this process and mathematical reasoning performed entirely in the latent space.
To this end, we investigate whether we can predict useful latent representations of the mathematical formulas that result from proof steps.
Ideally, we could rely entirely on predicted latent representations to sketch out proofs and only go back to the concrete mathematical formulas to check if our intuitive reasoning was correct.
This would allow for more flexible and robust system designs for automated reasoning. 

In this work, we present a first experiment indicating directly that theorem proving in the latent space might be possible. We build on HOList, an environment and benchmark for automated theorem provers based on deep learning~\cite{bansal2019holist} which is makes use of the interactive theorem prover HOL Light~\cite{Harrison96}, an interactive proof assistant. The HOList theorem database comprises of over 19 thousand theorems and lemmas from a variety of mathematical domains, including topology, multivariate calculus, real and complex analysis, geometric algebra, and measure theory. Concrete examples include basic properties of real and complex numbers such as $(x^{-1} = y) \Leftrightarrow (x = y^{-1})$, and also well-known theorems, such as Pythagoras' theorem, Skolem's theorem, the fundamental theorem of calculus, Abel's theorem for complex power series and that the eigenvalues of a complex matrix are the roots of its characteristic polynomial.

We focus on \emph{rewrite rules} (or \emph{rewrites} in short).
Rewrites are only one of several proof tactics in HOL Light, but they enable powerful transformations on mathematical formulas, as they can be given arbitrary theorems as parameters.
For example, the formula $3^2=z$ can be rewritten to $3\cdot 3=z$ by performing a rewrite with the parameter $x^2=x\cdot x$.
Alternatively, a rewrite may diverge (as it operates recursively) or it may return the same formula -- in both these cases we consider the rewrite to \emph{fail}.
For instance, in the example above, the rewrite would fail if we used equation $x+y = y+x$ as a rewrite parameter instead, since the expression $3^2=z$ does not contain any $+$ operators to match with.



In our experiments, we first train a neural network to map mathematical formulas into a latent space of fixed dimension. This network is trained by predicting -- based on the latent representation being trained -- whether a given rewrite is going to succeed (i.e. returns with a new formula). For successful rewrites we also predict the latent representation of the resulting formula.
To evaluate the feasibility of reasoning in latent space over two steps, we first predict the latent representation of the result of a rewrite, then we evaluate whether the predicted latent representation still allows for accurate predictions of the rewrite success of the resulting formula.
For multi-step reasoning beyond two steps, we predict the future latent representations based on the previous latent representation only - without seeing the intermediate formula.
Our experiments suggest that even after 4 steps of reasoning purely in latent space, neural networks show non-trivial reasoning capabilities, despite not being trained on this task directly.


\section{Related Work}

Our work is motivated by deep learning based automated theorem proving, but is also closely connected to model based reinforcement learning and approaches that learn to predict the future as part of reinforcement learning.

Model based reinforcement learning is concerned with creating a model of the environment while maximizing the expected reward (e.g.~\cite{ha2018recurrent}).
Already early works have shown that predicting the latent representations of reinforcement learning environments with deep learning is sometimes feasible - even over many steps~\cite{oh2015action,DBLP:journals/corr/ChiappaRWM17}.
This can enable faster training, since it can preempt the need for performing expensive simulations of the environment.
Predicting latent representation was also proposed in~\cite{brunner2018using} as a regularization method for reinforcement learning.

One recent successful example of model based reinforcement learning is~\cite{kaiser2019model}, where the system learns to predict the pixel-wise output of the Atari machine. However this approach is based on actually simulating the environment directly in the ``pixel space'' as opposed to performing predictions in a low dimensional semantic embedding space.
More related to our work is~\cite{piotrowskican}, which attempts to learn to rewrite simple formulas. The goal is there again is to predict the actual outcome of the rewrite rather than a latent representation of it. In~\cite{dosovitskiy2016learning}, they predict ``expected measurements'' as an extra supervision in addition to the reward signal.


Graph neural networks have been used for premise selection in higher order logic~\cite{wang2017premise} and more recently by~\cite{paliwal2019graph} as the core component of DeepHOL, a neural theorem prover for higher order logic. In this work, we build upon their neural network architecture, but utilize it for a different task.

\section{HOL Light}
\label{hollight}

HOL Light~\cite{Harrison96} is an {\it interactive proof assistant} (or interactive theorem prover) for higher-order logic reasoning. 
Traditionally, proof assistants have been used by human users for creating formalized proofs of mathematical statements manually. 
Although they come with limited forms of automation, it is still a cumbersome process to formalize proofs, even when it is already available in natural language.
Some large scale formalization efforts were conducted successfully in HOL Light and Coq~\cite{coq}, for example the formal proofs of the Kepler conjecture~\cite{hales2017formal} and that of the four color theorem~\cite{gonthier2008formal}. They required significant meticulous manual work and expert knowledge of the system.

Lately, there have been several attempts to improve the automation of the proof assistants significantly by so called ``hammers''~\cite{kaliszyk2015hol}. Still, traditional proof automation lacks the mathematical intuition of human mathematicians who can perform complicated intuitive arguments. The quest for modelling and automating fuzzy, ``human style'' reasoning is one of the main motivation for this work.


\subsection{Rewrite Tactic in HOL Light}
\label{rewrite}

The HOL Light system allows the user to specify a \emph{goal} to prove, and then offers a number of \emph{tactics} to apply to the goal. 
A tactic application consumes the goal and returns a list of subgoals.
Proving all of the subgoals is equivalent to proving the goal itself. 
Accordingly, if a tactic application returns the empty list of subgoals, the parent goal is proven.

In this work, we focus on the rewrite tactic (${\tt REWRITE\_TAC}$) of HOL Light, which is a particularly common and versatile tactic.
It takes a list of theorems as parameters (though in this work we only consider applications of rewrite with a single parameter).
Parameters must be an equation or a conjunction of equations; possibly guarded by a condition.
Given a goal statement $T$ and parameter $P_i$, the rewrite tactic searches for subexpressions in $T$ that match the left side of one of the equations in $P_i$ and replaces it with the right side of the equation.
The matching process takes care of variable names and types, such that minor differences can be bridged.
The rewrite procedure is recursive, and hence tries to rewrite the formula until no opportunities for rewrites are left.
The rewrite tactic also has a set of built-in rewrite rules, representing trivial simplifications, such as $\textrm{FST}(x,y)=x$.
Note that ${\tt REWRITE\_TAC}$ uses ``big step semantics'', meaning that the application of each individual operation can perform multiple elementary rewrite steps recursively.
For more details on  ${\tt REWRITE\_TAC}$, refer to the manual~\cite{rewrite-page}.

\section{Reasoning in Latent Space}

We embed higher-order logic statements into a fixed dimensional embedding space by applying a graph neural network to a suitably chosen graph representation of the corresponding formula.
The embedding is trained on predicting the outcome (success or failure) of a large number of possible formula rewrite operations.
Note that formulas can be quite complex as they are arbitrary typed lambda expressions in higher order logic.

For technical reasons, we will distinguish between two latent embedding spaces $L=\mathbb{R}^k$ and $L'=\mathbb{R}^k$ ($k=1024$) corresponding to two distinct embeddings for each formula, learned by two different networks.

We have trained three different models. $S$ denotes the set of syntactically correct higher-order logic formulas in HOL Light.
\begin{enumerate}
\item Rewrite success prediction $\sigma: S\times S\longrightarrow [0, 1] $,
\item Rewrite outcome prediction $\omega: S\times S\longrightarrow [0, 1] \times L$,
\item Embedding alignment prediction $\alpha:L\longrightarrow L'$.
\end{enumerate}
These networks and their purposes are described in detail in later subsections. 
Alternatively, we could use a single fixed embedding space with a single model predicting its own future embedding on the rewritten statement. 
That network could be trained end-to-end and reach better performance without the need of aligning the embedding spaces, removing the need for $\alpha$.
Here, we opted for a more controlled setup that relies on a fixed embedding network $\sigma$, trained for the sole task of predicting whether rewriting statement $T$ by $P$ is successful. This way we can  rely on a fixed embedding method and run more detailed ablation analyses. Merging $\sigma$ and $\omega$, is left for future work.

\subsection{Training Data}

We start with the theorem database of the HOList environment~\cite{bansal2019holist}, which contains 19591 theorems in its theorem database, approximately ordered by increasing complexity.
This is split into 11655 training, 3668 validation, and 3620 testing theorems.
To generate our training data, we generate all pairs $T, P$ of theorems from the training set, where $P$ must occur before $T$ in the database (to avoid circular dependencies). We then interpret theorem $T$ as a goal to prove and try to rewrite $T$ with $P$ using the ${\tt REWRITE\_TAC}$ of HOL Light. 
This can result in three different outcomes:
\begin{enumerate}
    \item $T$ is rewritten by theorem $P$ successfully, and the result differs from $T$.
    \item The rewrite operation terminates, but failed to change the input theorem $T$.
    \item The rewrite operation times out or diverges (becomes too big).
\end{enumerate}
In our experiments, we consider only the first outcome as successful, i.e.\ when the application finishes within the specified time limit and changes the target, as a successful rewrite attempt.
Each training example, therefore, consists of the pair $(T, P)$, the success/fail-bit of the rewrite (1 for successful rewrites, 0 for failed rewrites), and, for successful tactic applications, the new formula that results from the rewrite, which we denote with $R(T, P)$.


\subsection{Base Model Architecture and Training Methodology}
\begin{wrapfigure}{h}{0.3\textwidth}
  \centering
  \includegraphics[width=.3\textwidth]{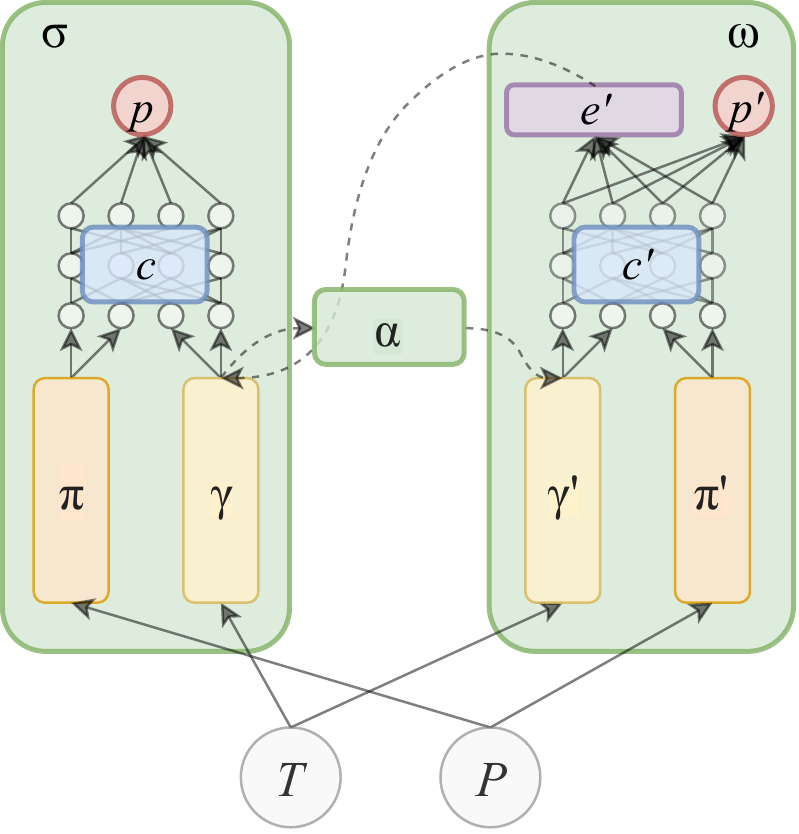}
  \vspace{-0.28 in}
  \caption{\small Depiction of network architecture}
  \label{fig:parameter_usage}
\end{wrapfigure}
The rewrite success prediction model $\sigma(T, P)$ is trained on the training set of theorems in the HOList benchmark. The training task is to predict the success or failure of the ${\tt REWRITE\_TAC}$ application.

We used a two-tower network (without weight sharing) with embedding towers $\gamma:S\longrightarrow L$ and $\pi$, one for each of the two formulas $T$ and $P$. Both towers are graph neural networks as described in~\cite{paliwal2019graph}.
Both of them embed the supplied formula in a fixed dimensional embedding space $\mathbb{R}^k$.
The concatenated embeddings are then processed by a three-layer perceptron $c$ with 
rectified linear activation, which is followed by $p$, a single output linear function predicting the logit and is trained by logistic regression on the success/fail-†bit of the rewrite. Formally: $\sigma(T, P)=p(c(\gamma(T), \pi(P)))$. 

\subsection{Outcome Prediction Model}
In addition to our base model, we train $\omega(T, P)$. 
This model has an identical two-tower embedding architecture as $\sigma$, but with a larger combiner network and an extra prediction layer $e'$ to predict embedding vector of the outcome of the rewritten formulas. Here the embedding towers are denoted by $\gamma'$ and $\pi'$, the combiner network is $c'$ and the two linear prediction layers are $p'$ and $e'$. That is: $\omega(T, P)=(p'(c'(\gamma'(T), \pi'(P))), e'(c'(\gamma'(T), \pi'(P)))$.

This model predicts both the success or failure of applying ${\tt REWRITE\_TAC}$ and for successful rewrites, the latent representation $\gamma(R(T,P))$ of the result.
While $p'$ is trained to predict the success of rewriting $T$ by $P$ by logistic regression, $e'$ is trained to predict $\gamma(R(T, P))$ by minimizing the squared error.

\subsection{Embedding Alignment Model}
Since $\omega$ and $\sigma$ produce latent vectors $\gamma(T)\in L$ and $\gamma'(T)\in L'$ in different spaces, we need to align those spaces enable deduction purely in the embedding space. (Merging $\sigma$ and $\omega$ will remove the need for the $\alpha$, however in our current setup we keep the embedding and deduction components separate). 

Given an initial statement $T$, we predict the approximate value of $\gamma(R(T, P))$, but in order to reason multiple steps in the embedding space alone, without explicitly computing $R(T, P)$, we need $\gamma'(R(T, P))$ to compute the outcome prediction of $\omega$. For that, we train a translation model $\alpha:L\longrightarrow L'$ which predicts $\gamma'(T)$ given an approximation of $\gamma(T)$. Note that $\alpha$ does not see $T\in S$ as an input, it makes its prediction based on the latent space representation $\gamma(T)$ of $T$ alone. This allows us for reasoning multiple steps ahead in the latent space without constructing any of the intermediate formulas explicitly.

\subsection{Reasoning}
\label{reasoning}
After we have trained our three models on the training set theorems (and theorem pairs) of the HOList benchmark, we can use them to perform rewrites in the latent space alone. 

We use $\sigma: S\times S\longrightarrow [0, 1]$ as a quality metric for the propagated embedding vector.
$\sigma(T, P)=p(c(\gamma(T), \pi(P)))$ was trained for predicting whether theorem $P$ rewrites $T$. Given an approximation $\tilde{v}_T$ of the latent representation $\gamma(T)$, we can evaluate $\tilde\sigma_T(P)$ defined by $\tilde\sigma_T(P)=p(\tilde{v}_T,\pi(P))$ for a large number of tactic parameters $P$. This is compared with true rewrite successes of $T$ by $P$ to assess the quality of the approximation.

To evaluate multiple steps of reasoning in the latent space, start with formula $T_1$ and rewrite by theorems $P_1,\cdots P_k$ in that order. For reasoning in the latent space, we only use approximate embeddings vectors of the resulting formulas. To assess the quality of the overall reasoning, the same sequence of rewrites is performed with the actual formulas and the final approximate embedding is evaluated with respect to the formula resulting from the sequence of formal rewrites.

In latent space $L=\mathbb{R}^k$ we from some initial theorem $T_1\in S$. The following schema indicates the sequence of predictions performed in latent spaces $L$ and $L'$:
$$T_1 \stackrel{\gamma'}{\longrightarrow} l_1'\in L' \stackrel{e'(\bullet,P_1)}{\longrightarrow} l_2\in L \stackrel{\alpha}{\longrightarrow} l_2' \in L' \stackrel{e'(\bullet, P_2)}{\longrightarrow} l_3 \in L \stackrel{\alpha}{\longrightarrow}l_3'\in L' \stackrel{e'(\bullet, P_3)}{\longrightarrow} \cdots $$
This way, we approximate the following sequence of deductions in the latent space $L$ alone, without producing any intermediate formulas. This is compared with the following formal sequence of rewrites:
$$T_1 \stackrel{R(\bullet, P_1)}{\longrightarrow} T_2 \stackrel{R(\bullet, P_1)}{\longrightarrow} T_3 \stackrel{R(\bullet, P_3)}{\longrightarrow} \cdots$$
That is, $l_2$ approximates the latent vector of $T_2$, that is $\gamma(R(T_1, P_1))$ and $l_3$ approximates the latent vector of $T_3$, that is $\gamma(R(R(T_1, P_1), P_2)$, by construction.

By a slight abuse of notation we will refer to the operation of one step of approximate deduction in the latent space by $\omega\circ\alpha$, as it is composition $\alpha$ and a subnetwork of $\omega$.


\section{Experiments}

This section provides an experimental evaluation that aims to answer the following question: Is this setup capable of predicting embedding vectors multiple steps ahead? We explore the prediction quality of the embedding vectors (for rewrite success) and see how the quality of predicted embedding vectors degrades  after $\omega\circ\alpha$ is used for predicting multiple steps in the latent space alone.

\subsection{Neural Network Architecture Details}

\begin{figure}{h}
  \centering
  \includegraphics[width=.6\textwidth]{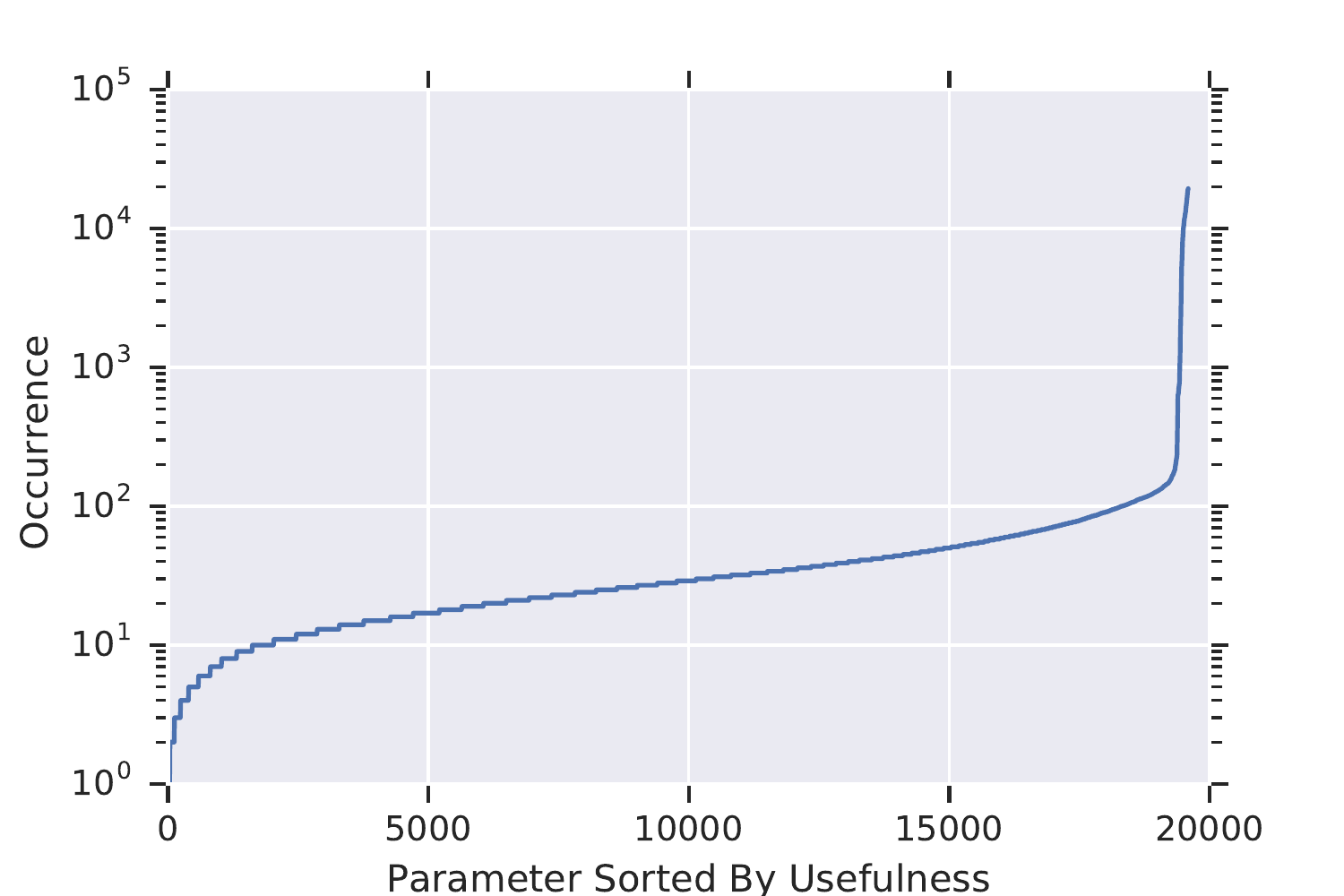}
  \caption{\small Graph showing the distribution of successful rewrites for each parameter $P$, computed over all $(T, P)$ pairs in the database. Note that a small portion of parameters can be used for many theorems. This allows the Random and Usage baselines to achieve above-chance performance by predicting success based on parameter alone. }
  \label{fig:parameter_usage}
\end{figure}

Our networks $\sigma$ and $\omega$ both have two towers, which are $16$-hop graph neural networks with internal node representations of $128$ dimensions. The output of each of the two towers is fed into a layer that expands the dimension of the node representation to $1024$ with a fully connected layer with shared weights for each node. This is followed by maximum pooling over all nodes of the network. The two resulting embedding vectors are concatenated along with their element-wise multiplication, and are processed by a three layer perceptron with rectified linear activation between the layers.

The same architecture is used for both $\sigma$ and $\omega$, but the two networks do not share weights. Also, $\omega$ has larger layers in its combiner network $c'$ than $\sigma$ in $c$, ($1024$ units each in $c'$ vs. $128$ units in the layers of $c$). This was necessary for producing good quality predictions of the embedding vector of the outcome of the rewrite, but unnecessary for predicting the rewrite success alone.

$\sigma$ and $\omega$ are trained with $g=16$ groups of $l+1=16$ instances in each batch: one successful example in each group and $l$ random negatives. However all other instances in other groups are used as negative instances for for each goal as well and they are considered negative regardless of whether they would rewrite it -- this is justified by the fact that only a few theorems rewrite any given $T_i$ so this introduces only a small amount of uncorrelated label noise. This training methodology is motivated by the fact that evaluating the combiner network is much cheaper then computing the embedding using the graph neural network.
Based on~\cite{alemi2016deepmath}, we expect that hard negative mining would improve our results significantly, but it is left for future work.

\subsection{Evaluation Dataset}
In order to measure the performance of our models after multiple deduction steps are performed, we generate datasets $D_0, D_1, \ldots, D_r$ successively by applying rewrites to a randomly selected set of examples. 
We start with all theorems from the validation set of HOList, denoted by $D_0$. We create $D_i$ from $D_{i-1}$ by selecting a random subset of $2000$ statements from the previous step and $200$ random tactic parameters ${\bf P}_T$ to rewrite by for each statement.
Formally, $D_i$ is defined by $\{R(T, P)\mid P\in{\bf P}_T\}$.

\subsection{Evaluation of Rewrite Prediction Model}
In order to evaluate the performance of $\sigma$ in isolation we need to compare it with carefully selected baselines:
\begin{enumerate}
    \item As Figure~\ref{fig:parameter_usage} shows, a few theorems are much more frequently applicable than others. We want to see how the prediction of rewrite success performs based on the rewrite parameter alone if we ignore the theorem to be rewritten. One way to establish such a baseline we just feed a randomly selected theorem $T'$ to $\sigma$ instead of $T$ to predict its rewrite success.
    \item A stronger ``baseline'' is achieved by utilizing the ground-truth to make the best prediction possible based on knowing $P$ but still being independent of $T$ (the theorem to be rewritten). This is the best achievable prediction that does not depend on $T$.
    \item As we have trained $\sigma$ and $\omega$ only on pairs of theorems from the original database, the models exhibit increasing generalization error as we evaluate them on formulas that with increasing number of rewrites. First, we measure the errors these models make if the theorem for the last step is evaluated directly. This gives an upper bound on the rewrite success prediction by $\omega$, since noisier embedding vectors end up with worse results on average.
    \item Finally, we want to measure how rate at which the latent vectors degrade as we propagate them in embedding space as described in Subsection~\ref{reasoning}.
\end{enumerate}
In order to measure the performance of our models after performing a given number of rewrite steps starting from the theorem database, we measure the tactic success prediction quality of $\sigma$ using predicted embeddings. To do so, we compute the ROC curve of the predictions and use the area under the curve (AUC) as our main metric. Higher curves and higher AUC values represent more accurate predictions of rewrite success. We measure how ROC curves change as we use different approximations of $\gamma(T)$.

\begin{figure}[h]
\centering
    \includegraphics[width=0.7\textwidth]{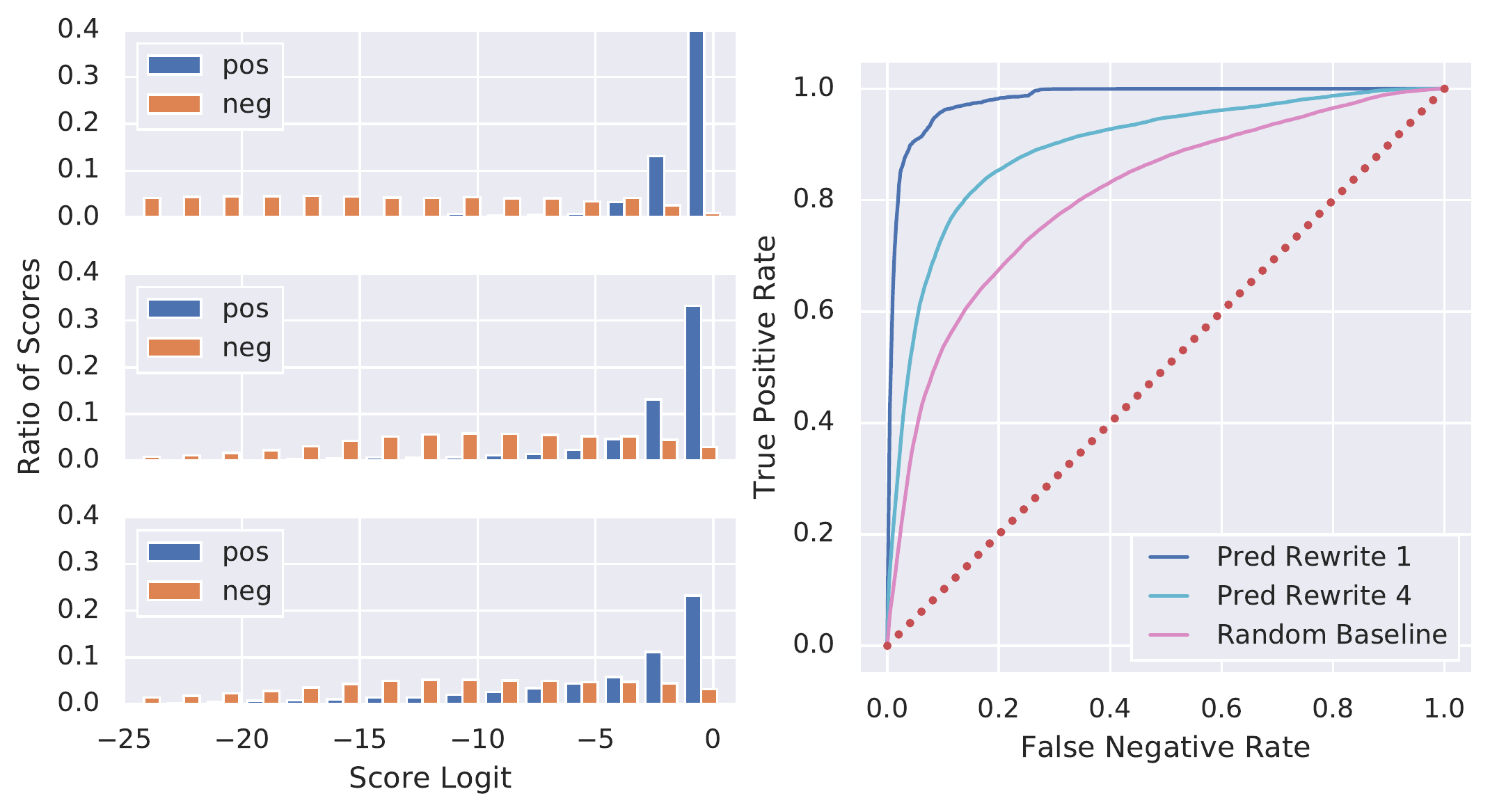}
    \caption{\small Left: Histograms of scores for first step of rewrite predictions (top), fourth step of rewrite predictions (middle), and random baseline (bottom). Right: Corresponding ROC curves. }
    \label{fig:histograms_vs_random}
\end{figure}

\begin{figure}[h]
    \centering
    \includegraphics[width=0.7\textwidth]{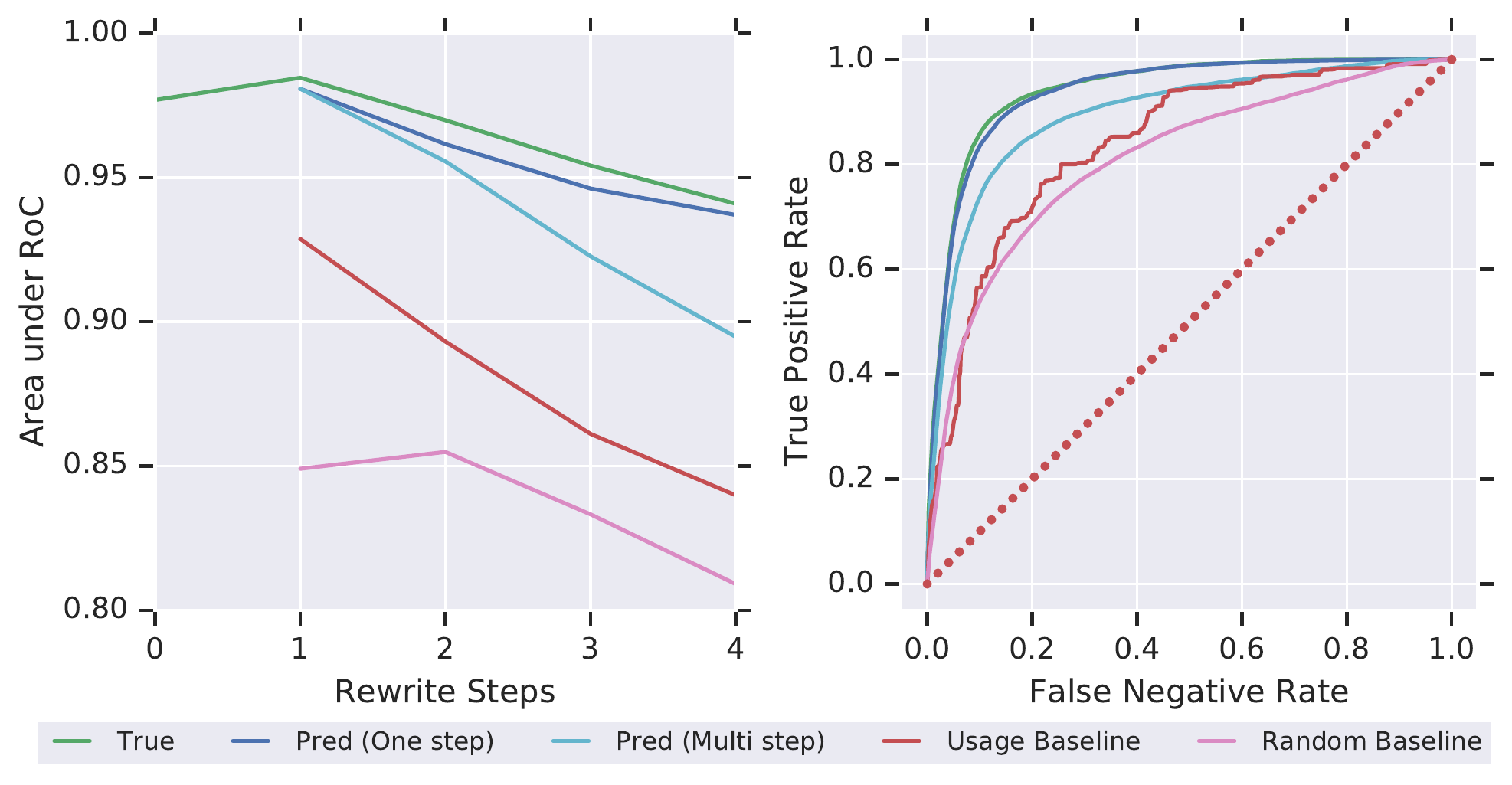}
    \caption{\small Left: Comparison of Area under ROC curves for each embedding method. Right: Comparison of ROC curves for each embedding method after the fourth rewrite.}
    \label{fig:auc_comparisons}
\end{figure}

\begin{figure}[h]
    \centering
    \includegraphics[width=0.7\textwidth]{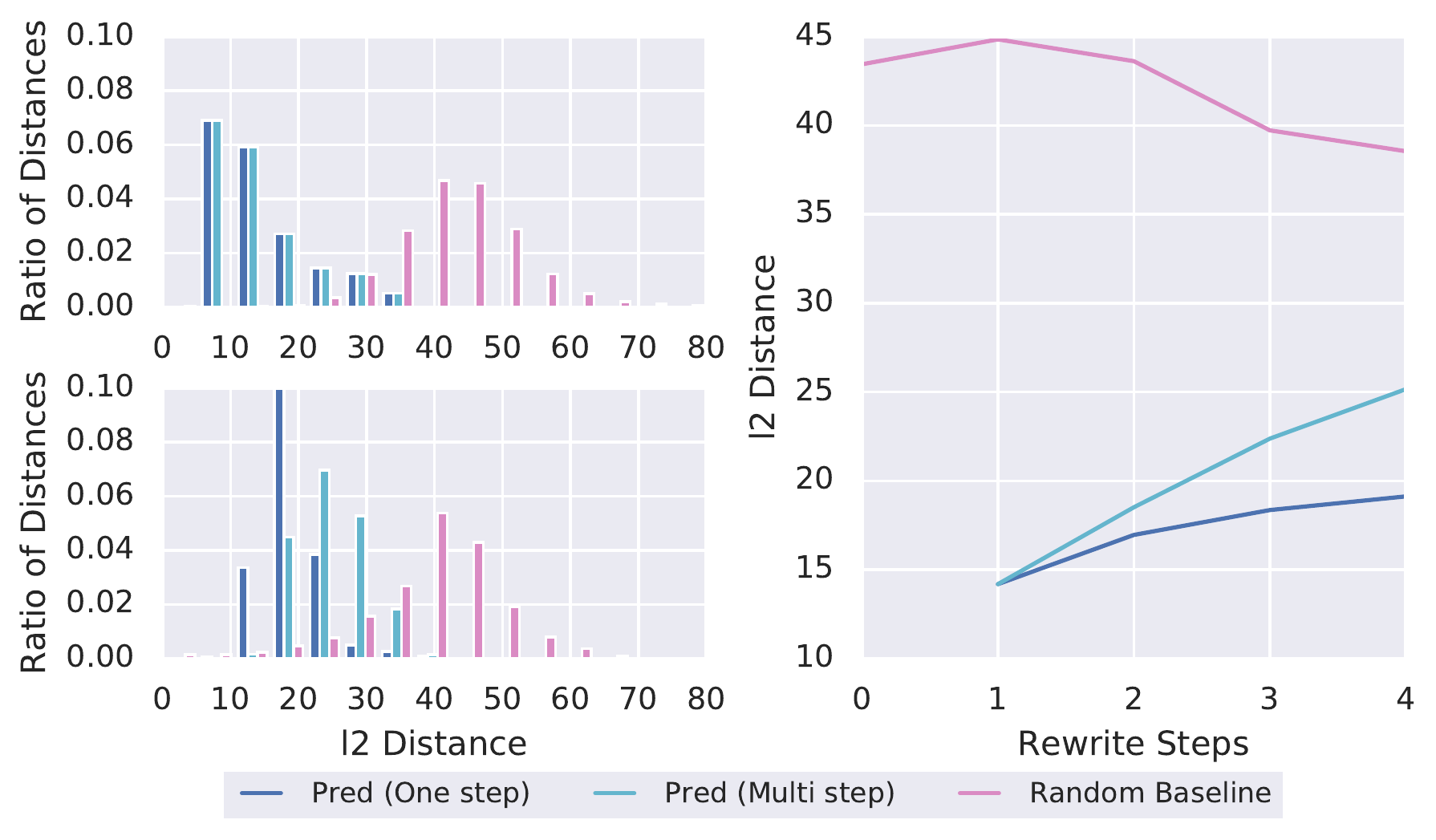}
    \vspace{-0.1in}
    \caption{\small Left: Histogram of $l^2$ distances between each embedding method and the true embedding for rewrite steps 1 and 4. Right: Plot showing the mean $l^2$ distances across rewrite steps.
    {\bf Pred (One step)}: expected error $\|\gamma(R(T,P)) - \alpha(e(\gamma'(T), \pi'(P)))\|_2$ between the true embedding of the rewritten statement $R(T,P)$ and that of the prediction network $\omega\circ\alpha$ one step ahead in $D_k$.
    {\bf Pred (Multi step)}: expected error for $\gamma(R(...R(T,P_1),...,P_k))$ in $D_k$ after $k$ approximate rewrites.
    {\bf Random baseline}: expected $l^2$ distance $\|\gamma(T_1)-\gamma(T_2)\|_2$ between the ``true'' embeddings of two randomly chosen theorems from $D_k$.}
    \label{fig:l2_distances}
\end{figure}

\begin{figure}[h]
    \begin{subfigure}[t]{0.41\textwidth}
        \centering
        \includegraphics[width=\textwidth]{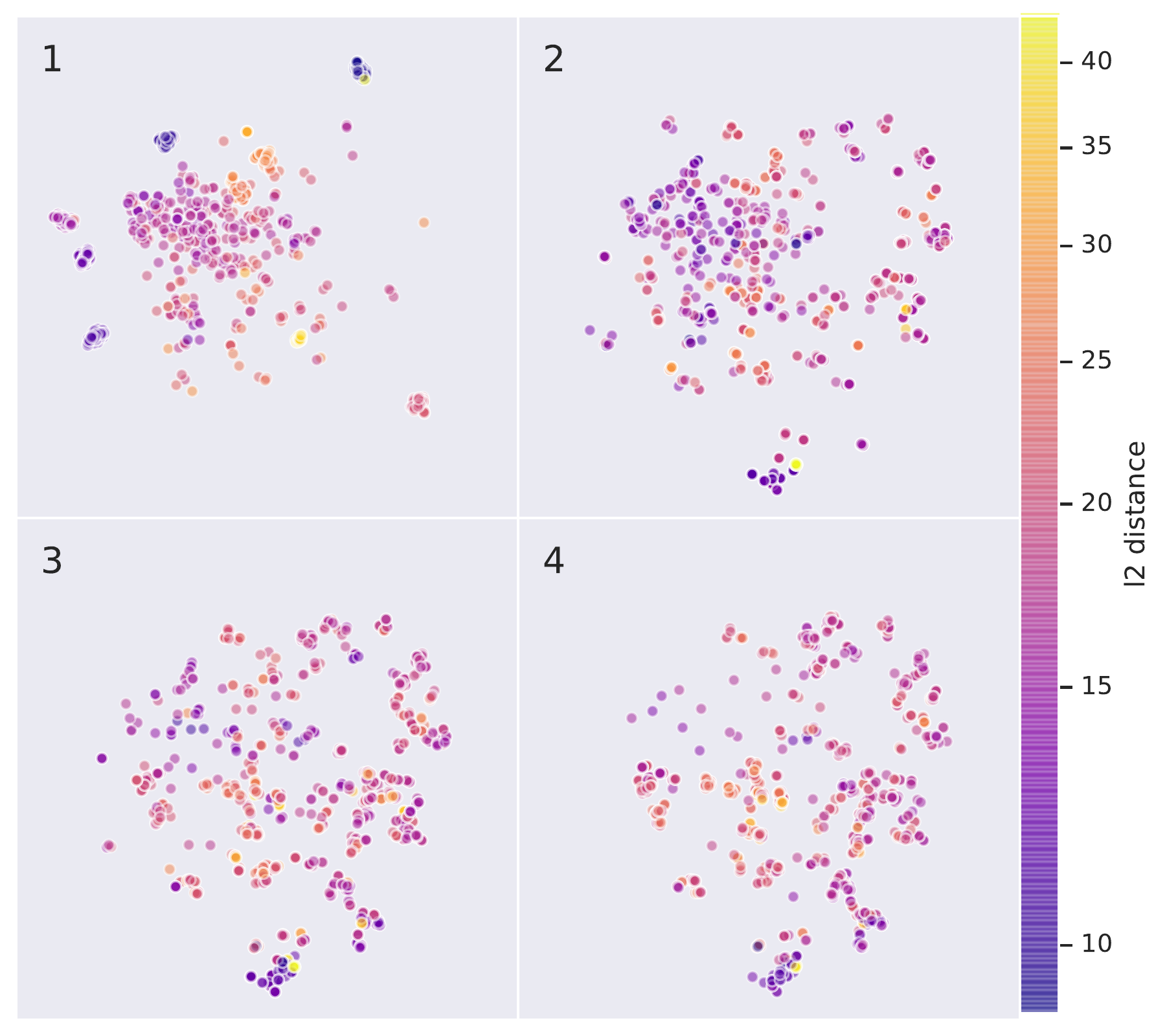}
        \caption{\small Visualization of the embedding spaces produced by embedding prediction $e^\prime$ in the latent space after $1,2,3$  and $4$ rewrite steps. The points are colored by their $l^2$-distance to the true embedding $\gamma(T)$.}
        \label{fig:embedding_spaces}
    \end{subfigure}
    \hspace{0.1in}
    \begin{subfigure}[t]{0.47\textwidth}
        \centering
        \includegraphics[width=\textwidth]{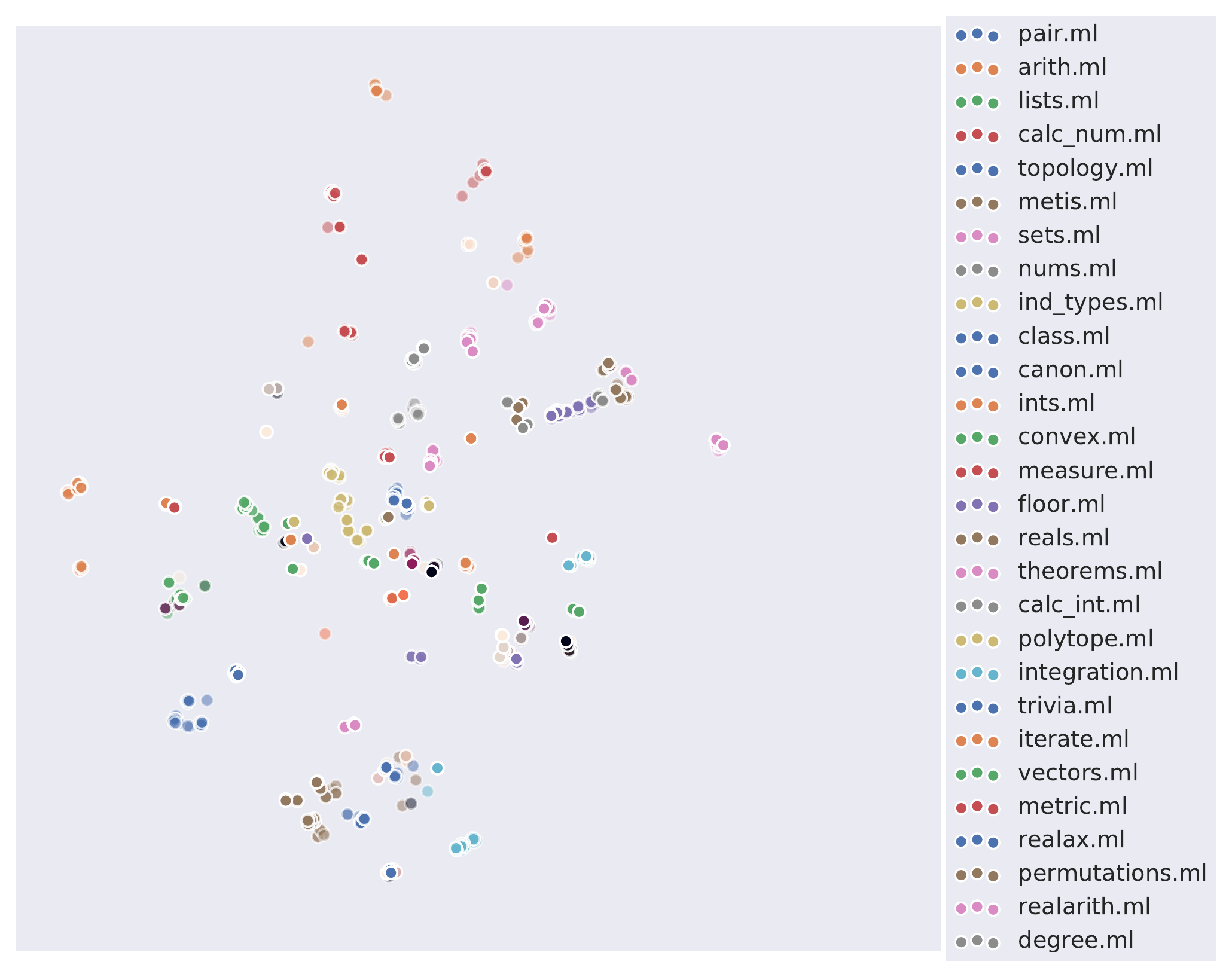}
        \caption{\small Visualization of the embedding space $\gamma(T)$ evolving over four rewrite steps. Theorems are colored by the area of mathematics they originate from, sourced from the theorem database. The brightness of the embeddings corresponds to the rewrite step in which it was generated, with more recent embeddings being darker. }
        \label{fig:theorems_by_color}
    \end{subfigure}
\end{figure}

\section{Analysis}

Figure~\ref{fig:histograms_vs_random} shows the distribution of the theorem pair prediction score logits of $p(c(\gamma(T), \pi(P)))$, for the those ``positive'' pairs that rewrite and the ``negative'' pairs that do not rewrite. Note that the ratio is normalized separately for the positive and negative pairs as negative pairs occur much more frequently than positive pairs.

One can see that the quality of the rewrite success prediction degrades significantly after four steps of reasoning purely in the latent space, but it is still much better than the random baseline.
This gives clear evidence that the embedding prediction manages to propagate much useful information over multiple steps in the latent space alone.

In Figure~\ref{fig:auc_comparisons} we make further measurements and comparisons on the quality of reasoning in the embedding space.
On the left hand side we measure five different metrics: 
The ``True'' curve assess the embedding $\gamma(T)$ computed directly from the target theorem $T$.
The ``Pred (One step)'' curve uses the approximate embedding $\alpha(e(\gamma'(T_{i-1}), \pi'(P_{i-1})))$ for $T_i=R(T_{i-1}, P_{i-1})$.
That is, we measure the degradation when performing a single step of embedding prediction.
The ``Pred (Multi-step)'' curve uses multiple steps of predictions completely in the latent space 
as described in Subsection~\ref{reasoning}. The ``Random Baseline'' predicts the rewrite success based on the latent vector of a random statement instead of the correct one.
``Usage Baseline'' is based on the constant prediction that ranks the parameters by how probably they rewrite any statement in the theorem database. This prediction is also independent of $T$.
One can see that our model could perform reasoning for $4$ steps in the embedding space and still retain a lot of the predictive power of the original model.

In order to appreciate the above results one should keep in mind that none one of our models $\alpha, \sigma$ and $\omega$ were trained on statements that were already rewritten. All the training was done only on the theorems present in the initial database. The reduction of prediction performance is apparent from downward trajectory of the ``Pred (Multi step)'' curve, which isolates this effect from that of the error accumulated by the embedding predictions, the effect of which is measured indirectly by the ``True'' curve.

In Figure~\ref{fig:l2_distances} we have measure the $l^2$ distance of the predicted embedding vectors versus that of the true embedding vectors $\gamma(T)$ of formulas after multiple rewrite steps in the latent space. These results are consistent with our earlier findings on success of rewrite predication after rewrite steps in the latent space: while there is some divergence of the predicted embedding vectors from the true embedding vectors (as computed from the rewritten statements directly), the predicted embedding vectors are significantly closer to the true embedding vectors than randomly selected embeddings.


\section{Conclusion}

In this paper we studied the feasibility of performing complex reasoning for mathematical formulas in a fixed $(1024)$ dimensional embedding space.
We proposed a new evaluation metric that measures the preservation semantic information under multiple reasoning steps in the embedding space.
Although our models were not trained for performing rewrites on rewritten statements, nor were they trained for being able to deduce multiple steps in the embedding space, our approximate rewrite prediction model $\omega\circ\alpha$ has demonstrated significant prediction power as far as $4$ approximate rewrite steps performed in the latent space.
Although it seems likely that these results can be significantly improved by better neural network architectures, hard negative mining and training on rewritten formulas,
our methods showcases a simple and efficient general methodology for reasoning in the latent space. In addition, it proposes an easy to use, fast to train and crisp evaluation methodology for representing mathematical statements by neural networks.

It is likely that such representations prove helpful for faster learning to prove without imitating human proofs like that in DeepHOL-Zero~\cite{bansal2019learning}, given that premise selection is a closely related task to predicting the rewrite success of statements. Self-supervised pre-training or even co-training such models with premise selection could prove useful as a way of learning more semantic feature representations of mathematical formulas.

\bibliography{bib}

\begin{thebibliography}{}

\bibitem[Alemi et~al., 2016]{alemi2016deepmath}
Alemi, A.~A., Chollet, F., Irving, G., E{\'e}n, N., Szegedy, C., and Urban, J.
  (2016).
\newblock Deepmath-deep sequence models for premise selection.
\newblock In {\em Advances in Neural Information Processing Systems}, pages
  2235--2243.

\bibitem[Bansal et~al., 2019a]{bansal2019learning}
Bansal, K., Loos, S.~M., Rabe, M.~N., and Szegedy, C. (2019a).
\newblock Learning to reason in large theories without imitation.
\newblock {\em arXiv preprint arXiv:1905.10501}.

\bibitem[Bansal et~al., 2019b]{bansal2019holist}
Bansal, K., Loos, S.~M., Rabe, M.~N., Szegedy, C., and Wilcox, S. (2019b).
\newblock {HOList}: An environment for machine learning of higher-order theorem
  proving.
\newblock {\em ICML 2019. International Conference on Machine Learning}.

\bibitem[Brunner et~al., 2018]{brunner2018using}
Brunner, G., Fritsche, M., Richter, O., and Wattenhofer, R. (2018).
\newblock Using state predictions for value regularization in curiosity driven
  deep reinforcement learning.
\newblock In {\em 2018 IEEE 30th International Conference on Tools with
  Artificial Intelligence (ICTAI)}, pages 25--29. IEEE.

\bibitem[Chiappa et~al., 2017]{DBLP:journals/corr/ChiappaRWM17}
Chiappa, S., Racani{\`{e}}re, S., Wierstra, D., and Mohamed, S. (2017).
\newblock Recurrent environment simulators.
\newblock {\em CoRR}, abs/1704.02254.

\bibitem[Coq, ]{coq}
Coq.
\newblock The {Coq Proof Assistant}.
\newblock \url{http://coq.inria.fr}.

\bibitem[Dosovitskiy and Koltun, 2017]{dosovitskiy2016learning}
Dosovitskiy, A. and Koltun, V. (2017).
\newblock Learning to act by predicting the future.
\newblock {\em ICLR 2017}.

\bibitem[Gauthier et~al., 2017]{gauthier2017tactictoe}
Gauthier, T., Kaliszyk, C., and Urban, J. (2017).
\newblock Tactictoe: Learning to reason with {HOL4} tactics.
\newblock In {\em LPAR-21. 21st International Conference on Logic for
  Programming, Artificial Intelligence and Reasoning}, volume~46, pages
  125--143.

\bibitem[Gonthier, 2008]{gonthier2008formal}
Gonthier, G. (2008).
\newblock Formal proof--the four-color theorem.
\newblock {\em Notices of the AMS}, 55(11):1382--1393.

\bibitem[Ha and Schmidhuber, 2018]{ha2018recurrent}
Ha, D. and Schmidhuber, J. (2018).
\newblock Recurrent world models facilitate policy evolution.
\newblock In Bengio, S., Wallach, H., Larochelle, H., Grauman, K.,
  Cesa-Bianchi, N., and Garnett, R., editors, {\em Advances in Neural
  Information Processing Systems 31}, pages 2450--2462. Curran Associates, Inc.

\bibitem[Hales et~al., 2017]{hales2017formal}
Hales, T., Adams, M., Bauer, G., Dang, T.~D., Harrison, J., Le~Truong, H.,
  Kaliszyk, C., Magron, V., McLaughlin, S., Nguyen, T.~T., et~al. (2017).
\newblock A formal proof of the {Kepler} conjecture.
\newblock In {\em Forum of Mathematics, Pi}, volume~5. Cambridge University
  Press.

\bibitem[Harrison, 1996]{Harrison96}
Harrison, J. (1996).
\newblock {HOL Light}: A tutorial introduction.
\newblock In {\em FMCAD}, pages 265--269.

\bibitem[HOL Light Rewrite Tactic Reference, ]{rewrite-page}
HOL Light Rewrite Tactic Reference.
\newblock Accessed: 2019/09/23.

\bibitem[Kaiser et~al., 2019]{kaiser2019model}
Kaiser, L., Babaeizadeh, M., Milos, P., Osinski, B., Campbell, R.~H.,
  Czechowski, K., Erhan, D., Finn, C., Kozakowski, P., Levine, S., et~al.
  (2019).
\newblock Model-based reinforcement learning for {Atari}.
\newblock {\em arXiv preprint arXiv:1903.00374}.

\bibitem[Kaliszyk and Urban, 2015]{kaliszyk2015hol}
Kaliszyk, C. and Urban, J. (2015).
\newblock Hol (y) hammer: Online atp service for hol light.
\newblock {\em Mathematics in Computer Science}, 9(1):5--22.

\bibitem[Lederman et~al., 2018]{Lederman2018QBF}
Lederman, G., Rabe, M.~N., and Seshia, S.~A. (2018).
\newblock Learning heuristics for automated reasoning through deep
  reinforcement learning.
\newblock {\em CoRR}, abs/1807.08058.

\bibitem[Loos et~al., 2017]{loos2017deep}
Loos, S., Irving, G., Szegedy, C., and Kaliszyk, C. (2017).
\newblock Deep network guided proof search.
\newblock {\em LPAR-21. 21st International Conference on Logic for Programming,
  Artificial Intelligence and Reasoning}.

\bibitem[Oh et~al., 2015]{oh2015action}
Oh, J., Guo, X., Lee, H., Lewis, R.~L., and Singh, S. (2015).
\newblock Action-conditional video prediction using deep networks in {Atari}
  games.
\newblock In {\em Advances in neural information processing systems}, pages
  2863--2871.

\bibitem[Paliwal et~al., 2019]{paliwal2019graph}
Paliwal, A., Loos, S., Rabe, M., Bansal, K., and Szegedy, C. (2019).
\newblock Graph representations for higher-order logic and theorem proving.
\newblock {\em arXiv preprint arXiv:1905.10006}.

\bibitem[Piotrowski et~al., 2019]{piotrowskican}
Piotrowski, B., Brown, C., Urban, J., and Kaliszyk, C. (2019).
\newblock Can neural networks learn symbolic rewriting?

\bibitem[Wang et~al., 2017]{wang2017premise}
Wang, M., Tang, Y., Wang, J., and Deng, J. (2017).
\newblock Premise selection for theorem proving by deep graph embedding.
\newblock In {\em Advances in Neural Information Processing Systems}, pages
  2786--2796.

\end{thebibliography}
\bibliographystyle{apalike}
 
\end{document}